\documentclass[11pt]{article}
\usepackage{amsmath}
\usepackage{amssymb}
\usepackage{float}
\usepackage{graphicx}

\usepackage[comma,sort&compress]{natbib}
\usepackage{color} 

\usepackage{lineno}
\usepackage{url}
\usepackage{pdfpages}
\usepackage{setspace} 
\usepackage{lineno}

\usepackage{hyperref}

\usepackage[table]{xcolor}

\hypersetup{
	bookmarks=true,         % show bookmarks bar?
	unicode=false,          % non-Latin characters in Acrobat‚????s bookmarks
	pdftoolbar=true,        % show Acrobat‚????s toolbar?
	pdfmenubar=true,        % show Acrobat‚????s menu?
	pdffitwindow=false,     % window fit to page when opened
	pdfstartview={FitH},    % fits the width of the page to the window
	pdftitle={My title},    % title
	pdfauthor={Author},     % author
	pdfsubject={Subject},   % subject of the document
	pdfcreator={Creator},   % creator of the document
	pdfproducer={Producer}, % producer of the document
	pdfkeywords={keywords}, % list of keywords
	pdfnewwindow=true,      % links in new window
	colorlinks=true,       % false: boxed links; true: colored links
	linkcolor=blue,          % color of internal links
	citecolor=blue,        % color of links to bibliography
	filecolor=magenta,      % color of file links
	urlcolor=cyan           % color of external links
}

\usepackage[labelfont=bf,labelsep=period,justification=raggedright]{caption}

\makeatletter
\renewcommand{\@biblabel}[1]{\quad#1.}
\makeatother

\date{}

\title{Understanding Emergent Behaviours in   Multi-Agent Systems with Evolutionary Game Theory}

\author{The Anh Han$^{1,\star}$}

\begin{document}
	\maketitle
	{\footnotesize
		\noindent
		$^{1}$ School of Computing, Engineering and Digital Technologies,  Teesside University, Middlesbrough, UK TS1 3BA\\
		$^\star$ Corresponding: The Anh Han (t.han@tees.ac.uk) 
	}

\newpage
\begin{abstract}
The mechanisms of emergence and evolution  of collective behaviours in dynamical Multi-Agent Systems (MAS) of multiple interacting agents, with diverse behavioral strategies in co-presence, have been undergoing mathematical study via Evolutionary Game Theory (EGT). Their systematic study also resorts to agent-based modelling and simulation (ABM)  techniques, thus enabling the study of aforesaid mechanisms under a variety of conditions, parameters, and alternative virtual games.
This paper summarises some main research directions and  challenges  tackled in our group,  using methods from EGT and ABM. These range from the introduction of cognitive and emotional mechanisms into agents' implementation in an evolving MAS, to the cost-efficient interference for promoting prosocial behaviours in complex networks,  to the regulation and governance of AI safety development ecology, and to the equilibrium analysis of random evolutionary multi-player games.  
This brief  aims to sensitize the reader to  EGT based issues, results and prospects, which are accruing in importance for the modeling of minds with machines and the engineering of prosocial behaviours in dynamical MAS, with impact on our understanding of the emergence and stability  of  collective behaviours.
In all cases, important open problems  in MAS research as viewed or prioritised by the group are described. \\


\end{abstract}

 \newpage

\section{Introduction}
The problem of promoting the emergence and stability of diverse collective behaviours in populations of abstract individuals have been undergoing mathematical study via Evolutionary Game Theory (EGT) \citep{nowak:2006bo,hardin:1968mm,axelrod:1984yo,sigmund:2010bo}. 
They have attracted sustained attention from fields as diverse as  Multi-Agent Systems (MAS), Economics, Biology and Social Sciences. 
%Their systematic study resorts to simulation techniques, thus enabling the study of aforesaid mechanisms under a variety of conditions, parameters, and alternative virtual games. The theoretical and experimental results have continually been surprising, rewarding, and promising.
%The problem of explaining (rigorously) the evolution of cooperative behaviour was described as one of the 125 most important questions faced by Sciences in the  Science journal \citep{pennisi2005did}.

Since its inception, EGT has become a powerful mathematical framework for the modelling and analysis of complex, dynamical  MAS %
\citep{hofbauer:1998mm,HanJaamas2016,paiva2018engineering}, in biological, social contexts as well as computerised systems \citep{pereira2021employing,tuyls2007evolutionary,HanBook2013,santos2020picky,perc2017statistical}. 
 It has  been used widely and successfully to study numerous important and challenging questions faced by many disciplines and societies,  such as:  what are the mechanisms underlying the evolution of cooperative behaviour at various levels of organisation (from genes to human society)  \citep{nowak:2006pw,perc2017statistical}? How to mitigate existential risks such as those posed by climate change  \citep{santos:2011pn,santos2020picky} or advanced Artificial Intelligence (AI) technologies \citep{han2019modelling}? What are the roles of cognition and emotions in behavioural evolution  \citep{Luis2017AAMAS, HanBook2013}? 

In EGT, the payoff agents obtained from interacting with other agents in the system,  is interpreted as individual fitness. An important property in EGT is the so called frequency dependent selection, where the fitness of an individual does not only depend on its strategy, but also on the composition of the population in relation with (multiple) other strategies, naturally leading to a dynamical approach.  
EGT was originated back in 1973 with John Maynard Smith and George R. Price's formalisation of animal contests, extending classical game theory  \citep{von2007theory} to provide the mathematical criteria that can be used to predict the evolutionary outcomes  of interaction among competing strategies \citep{SP73}.

Diverse mathematical approaches and methods for analysing EGT models have been developed over the years. These include continuous approaches such as replicator equations which assume large population limits, unusually  requiring analysis of  systems of differential equations \citep{hofbauer:1998mm}. In finite size systems,   stochastic approaches are required and usually approximations methods are needed (e.g. using mean-field analysis). Here computer simulations and methods from statistical physics, such as Monte Carlo simulations, are very useful to analyse highly complex systems where analytical results are hard to achieve, for example, when populations are distributed on complex networks \citep{perc2017statistical}. 

Resorting to these complementary approaches, we have been tackling a number of fundamental research challenges in MAS and collective behaviour studies. 
First, we initiated the introduction in an evolving MAS, of cognitive and emotional capabilities, as inspired on techniques and theories of AI, namely those pertaining to intention recognition, commitment, trust and guilt (Section \ref{Section:cognition}).  Second, we initiated  baseline EGT models that capture the complex ecology of choices associated with a competition for domain supremacy using AI technology, allowing ones to formally explore various regulatory proposals of AI  development (Section \ref{Section:AIRace}). Third, we introduced cost-efficient optimisation problems in dynamical MAS on complex networks in order to capture various stochastic aspects of MAS interactions and evolution  over time (Section \ref{Section:interference}). Finally, we described our extensive analysis of  statistical properties of the number of (stable) equilibria  in a random evolutionary multi-player game, providing insights on generic equilibrium properties of large, dynamical MAS (Section \ref{Section:randomgames}).

\section{Cognitive and Emotional Mechanisms for Promoting Prosocial Behaviours in dynamical MAS } 
\label{Section:cognition}
% \section{Cognitive, trust-based and emotional mechanisms for promoting cooperation }
% Intention recognition, anticipation (zisis et al), trust (cogn system paper), guilt (aamas), apology, commitment

%The problem of evolution of cooperation and of the emergence of collective action---cutting across areas as diverse as Biology, Economy, Artificial Intelligence, Political Science, or Psychology---is one of the greatest interdisciplinary challenges science faces today  \citep{hardin:1968mm,key:axelrod84,nowak:2006bo,key:Sigmund_selfishnes}. To understand the evolutionary mechanisms that promote and keep cooperative behaviour is all the more complex as increasingly intricate is the intrinsic complexity of the partaking individuals. 

In its simplest form, a cooperative act is metaphorically described as the act of paying a cost to convey a benefit to someone else. If two players simultaneously decide to cooperate or not, the best possible response will be to try to receive the benefit without paying the cost. In an evolutionary setting, we may also wonder why would natural selection equip selfish individuals with altruistic tendencies while it incites competition between individuals and thus apparently rewards only selfish behavior?  Several mechanisms responsible for promoting cooperative behavior have been recently identified, ranging from kin and group ties, to different forms of reciprocity and networked populations (see surveys in  \citep{sigmund:2010bo,nowak:2006pw,perc2017statistical}). 

Moreover, more complex strategies based on the evaluation of interactions between third parties allow the emergence of kinds of cooperation that are immune to exploitation because then interactions are channelled to just those who cooperate. Questions of justice and trust, with their negative (punishment) and positive (help) incentives, are fundamental in games with large diversified groups of individuals gifted with intention recognition capabilities. In allowing them to choose amongst distinct behaviours based on suggestive information about the intentions of their interaction partners---these in turn influenced by the behaviour of the individual himself---individuals are also influenced by their tolerance to error or noise in the communication. One hopes that, to start with, understanding these capabilities can be transformed into mechanisms for spontaneous organization and control of swarms of autonomous robotic agents {\citep{bonabeau},  these being envisaged as large populations of agents where cooperation can emerge, but not necessarily to solve a priory given goals, as in distributed MAS.

\paragraph{Intention recognition}
The ability  of recognizing (or reading)  intentions of others has been observed and shown to play an important role  in many cooperative interactions, both in humans and primates \citep{key:tomassello2008,key:Meltzoff2005}. 
Intentions and their recognition play a central role in morality, as shown for example in studies on the doctrines of double and triple effect \citep{ha07a-moral}. 
However, most studies on the evolution of cooperation, grounded on evolutionary dynamics and game theory, have neglected the important role played by a basic form of intention recognition in behavioral evolution. In our work  \citep{key:hanijcai2011,key:hanetalAdaptiveBeh,key:HanetalAlife,han2015synergy}, we have addressed explicitly this issue, characterizing the dynamics emerging from a population of intention recognizers.

Intention recognition was implemented using Bayesian Network probabilistic methods, taking into account the information of current signals of intent, as well as the mutual trust and tolerance accumulated from previous one-on-one play experience -- including how an agent's previous defections may influence another agent's intent -- but without resorting to information gathered regarding agents' overall reputation in the population. A player's present intent can be understood here as how he is going to behave in the next round, whether by cooperating or defecting. Intention recognition can also be learned from a corpus of prior interactions among game strategies, where each strategy can be envisaged and detected as players' (possibly changing) intent to behave in a certain way \citep{HanBook2013}. In both cases, we experimented with populations with different proportions of diverse strategies in order to calculate, in particular, what is the minimum fraction of individuals capable of intention recognition for cooperation to emerge, invade, prevail, and persist. 

It is noteworthy that intention recognition techniques have been studied actively in AI for several decades \citep{key:charniak93,key:sadri2010_logicbasedIR,hanpereiraAICOM2013,sukthankar2014plan}, with several applications such as for improving human-computer interactions, assisting living, moral reasoning,  and team work \citep{key:roy2007,HanChapterBehRecog2013}. In most of these applications the agents engage in repeated interactions with each other. Our results suggest that equipping the agents with an ability to recognize intentions of others can improve their cooperation and reduce misunderstanding that can result from noise and mistakes.

The notions of intentions used in our  two models have been specialized for the concrete game-theoretical contexts in place. A more general definition, e.g. as described in Bratman's seminal work  \citep{key:bratman1987}, can accommodate for intention changes or even abandonment. For instance, a player can change his intention or strategy before the next interaction or round of game takes place. That aspect was  not considered in our EGT models, as players can change their strategy only at the end of a generation \citep{sigmund:2010aa}. We envisage a convenient extension to that direction since  our intention recognition methods---performed through Bayesian Network inference techniques, as  described in \citep{han2013context}---can cope with intention changes and abandonment. %even more, they have been tested with a benchmark generated in the EGT context itself \citep{key:hanluis_uai_workshop_2011}.  Further discussion of the intention recognition models in the EGT context and possible extensions can be found in \citep[Chapter 6]{HanBook2013}.   

\paragraph{Commitments}
Agents make commitments towards others when they give up options in order to influence others. Most commitments depend on some incentive that is necessary to ensure that an action (or even an intention) is in the agent's interest and thus will be carried out in the future \citep{nesse2001evolution,gintis:2000bv}. Asking for prior commitments can just be used as a strategy to clarify the intentions of others, whilst at the same time manifesting our own \citep{han2015synergy}. 
All parties then clearly know to what they commit and can refuse such a commitment whenever the offer is made \citep{HanJaamas2016}. 

In our work \citep{han2013good,key:Hanetall_AAMAS2012,HanJaamas2016,HanInterface2022},  we investigate analytically and numerically whether costly commitment strategies, in which players propose, initiate and honor a deal, are viable strategies for the evolution of cooperative behavior. 
Resorting the EGT mathematical analysis and simulations, we have shown that when the cost of arranging a commitment is justified with respect to the benefit of cooperation, substantial levels of cooperation can be achieved, especially when one insists on sharing the arrangement cost  \citep{han2013good}. On the one hand, such commitment proposers can get rid of fake committers  by proposing a strong enough compensation cost. On the other hand, they can maintain a sufficient advantage over the commitment free-riders, because a commitment proposer will cooperate with players alike herself, while the latter defect among themselves. We have also compared the commitment  strategy with the simple costly punishment strategy---an important one for the evolution of cooperation \citep{key:fehr2002}---where no prior agreements are made. The results show that the first strategy leads to a higher level of cooperation than the latter one. Furthermore, we have shown that these observations regarding evolutionary viability of commitment-based strategies are also applicable to other collective behaviours including coordination \citep{ogbo2021evolution} and AI safety \citep{han2022voluntary}.

It is noteworthy that there has been an extensive literature of AI and MAS  research on commitment, e.g.,  \citep{Chopra09m.p.:multiagent,Wooldridge99,castelfranchi2010,singh2013norms}, but the main concern therein is how to formalize different aspects  of commitment and how a commitment mechanism  can be implemented in multi-agent interactions to enhance these (e.g. for improved collaborative problem solving \citep{Wooldridge99}), especially in the context of classical game theory. Our work would provide insights into the design of MAS that rely on commitments or punishment in order to promote  collective behaviour among agents \citep{hasan2013emergence}.

%Note that in our work we specify  commitments in the context of one-shot  interactions.  When considering repeated games, such as the repeated PD, we might extend the specification by having conditional commitments, such as conditional promises and threats \citep{Schelling1960}. It would be interesting to ask in this context whether a  long-term relationship is better sustained through direct reciprocity or through proposing long-term commitments, and how those two mechanisms interact. 

\paragraph{Trust in autonomous systems (hybrid MAS)}
The actions of intelligent agents, such as chatbots, recommender systems, and virtual assistants are typically not fully transparent to the user \citep{Beldad:2016:a}. Consequently, users take the risk that such agents act in ways opposed to the users' preferences or goals \citep{Luhmann:1979:a,DhakalRSOS2022}. It is often argued that people use trust as a cognitive shortcut to reduce the complexity of such interactions. 

In our recent work \citep{HanetalTrust2021},  we study  this by using EGT modelling to examine  the viability of  trust-based strategies in the context of an iterated prisoner's dilemma  game \citep{sigmund:2010aa}. We show that these strategies can  reduce the opportunity cost of verifying whether the action of their co-player was actually cooperative, and out-compete strategies that are always conditional, such as Tit-for-Tat. We argue that the opportunity cost of checking the action of the co-player is likely to be greater when the interaction is between people and intelligent artificial agents, because of the reduced transparency of the agent.

 In our work, trust-based strategies  are reciprocal strategies that cooperate as long as the other player is observed to be cooperating. Unlike classic reciprocal strategies, once mutual cooperation has been observed for a threshold number of rounds they stop checking their co-player's behaviour every round, and instead only check it with some probability. By doing so, they reduce the opportunity cost of verifying whether the action of their co-player was actually cooperative. We demonstrate that these trust-based strategies can out-compete strategies that are always conditional, such as Tit-for-Tat, when the opportunity cost is non-negligible.

We argue that this cost is likely to be greater when the interaction is between people and intelligent agents within a hybrid MAS, since the interaction becomes less transparent to the user (e.g. when it is done over the internet), and artificial agents have limited capacity to explain their actions compared to humans \citep{Pu:2007:a}. 
Consequently, we expect people to use trust-based strategies more frequently in interactions with intelligent agents. Our results provide new, important insights into the design of mechanisms for facilitating interactions between humans and intelligent agents, where trust is an essential factor.

Trust is a commonly observed mechanism in human interactions, and discussions on the role of trust are being extended to social interactions between humans and intelligent machines \citep{Andras2018TrustingSystems}. It is therefore important to understand how people behave when interacting with those machines; particularly, whether and when they might exhibit trust behaviour towards them? Answering this is crucial for designing mechanisms to facilitate human-intelligent machine interactions, e.g. in engineering pro-sociality in a hybrid society of humans and machines \citep{paiva2018engineering}.

\paragraph{Guilt}
Machine ethics involving the capacity for artificial intelligence to act morally is an open project for scientists and engineers \citep{pereira2020machine}. One important concern is how to represent emotions that are thought to modulate human moral behaviour, such as guilt, in computational models. Upon introspection, guilt is present as a feeling of being worthy of blame for a moral offence. Burdened with guilt, an agent may then act to restore a blameless internal state in which this painful emotion is no longer present.

Inspired by psychological and evolutionary studies, we have constructed an EGT model representing guilt in order to study its role in promoting pro-social behaviour \citep{Luis2017AAMAS}. We modelled guilt in terms of two characteristics. First, guilt involves a record of transgressions which is formalised by the number of offences. Second, guilt involves a threshold over which the guilty agent must alleviate the associated emotional pain, in the case of our models, through apology \citep{ijcai2013TAH}, and also by involving self-punishment \citep{Luis2017AAMAS}, as required by the guilty feelings, both of which affect the payoff for the guilty agent. With this work, we were able to show that cooperation does not emerge when agents alleviate guilt without considering their co-players’ attitudes about the alleviation of guilt too. In that case, guilt-prone agents are easily dominated by agents who do not express guilt or who are without motivation to alleviate their own guilt. When, on the other hand, the tendency to alleviate guilt is mutual, and the guilt-burdened agent alleviates guilt in interactions with co-players who also act to alleviate guilt when similarly burdened, then cooperation thrives.

From a MAS perspective, including mixed social-technological communities encompassing potentially autonomous artificial agents, and invoking the so-called “value alignment” problem \citep{gabriel2020artificial}, our models confirm that conflicts can be avoided when morally salient emotions, like guilt, help to guide participants toward acceptable behaviours. In this context, systems involving possible future artificial moral agents may be designed to include guilt, so as to align agent-level behaviour with human expectations, thereby resulting in overall social benefits through improved cooperation, as evinced by our prospective work on modelling guilt.

\section{Governance of Artificial Intelligence development: A Game-Theoretical  Approach}
\label{Section:AIRace}
%\tr{Challenge/big picture: when a race emerges; cooperation/collaboration emerges; mapping AI scenarios and governance }
Rapid technological advancements in Artificial Intelligence (AI), together with the growing deployment of AI in new application domains such as robotics, face recognition, self-driving cars, genetics, are generating an anxiety which makes companies, nations and regions think they should respond competitively \citep{cave2018ai,lee2018ai}. AI appears for instance to have instigated a race among chip builders, simply because of the requirements it imposes on the technology.  Governments are furthermore stimulating economic investments in AI research and development as they fear of missing out, resulting in a racing narrative that increases further the anxiety among stake-holders \citep{cave2018ai}. 

Races for supremacy in a domain through AI may however have detrimental consequences since participants to the race may well ignore ethical and safety checks in order to speed up the development and reach the market first. AI researchers and governance bodies, such as the EU, are urging to consider together both the normative and the social impact of major technological advancements concerned \citep{EUAIwhitepaper2020}. However, given the breadth and depth of AI and its advances, it is not an easy task to assess when and which AI technology in a concrete domain needs to be regulated. This issue was, among others, highlighted in the recent EU White Paper on AI \citep{EUAIwhitepaper2020} and the UK  National AI strategy.  %Data to estimate the risk of a technology is usually limited, especially at an early stage of its development or deployment. 

Several proposals for mechanisms on how to avoid, mediate, or regulate the development and deployment of  AI  \citep{cave2018ai,o2020windfall,cimpeanu2022artificial}. Essentially, regulatory measures such as restrictions and incentives are proposed to limit harmful and risky practices in order to promote beneficial designs \citep{baum2017promotion}.  Examples include financially supporting the research into beneficial AI and  making AI companies pay
fines when found liable for the consequences of harmful AI. 

\begin{figure*}
\centering
\includegraphics[width=0.9\linewidth]{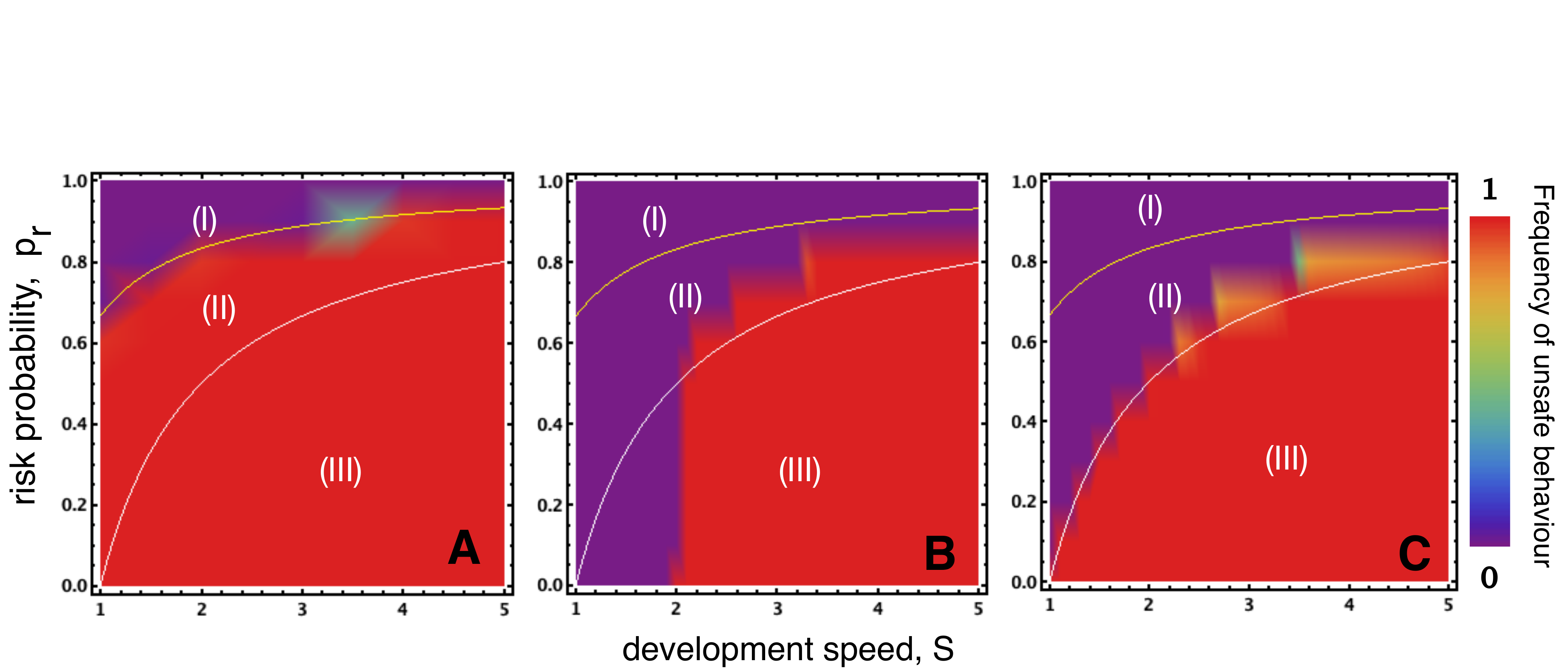} 
\caption{Frequency of unsafe behaviour as a function of development speed ($s$) and the disaster risk ($p_r$). \textbf{Panel A,  in absence of incentives} \citep{han2019modelling}, the parameter space can be split into three regions. In regions (I) and (III), safe and unsafe/innovation, respectively, are the preferred collective outcome also selected by natural selection, thus no regulation being required. Region (II) requires regulation as safe behaviour is preferred but not the one selected. \textbf{Panel B,  when unsafe behaviour is sanctioned unconditionally} \citep{han2020Incentive}, while unsafe behaviour is reduced in region II, over-regulation occurs in region III, reducing beneficial innovation. \textbf{Panel C, unsafe behaviour is sanctioned only in presence of a voluntary commitment} \citep{han2022voluntary}, unsafe behaviour is significantly reduced in region II while avoiding over-regulation.  
}
\label{fig}
\end{figure*}

Although such regulatory measures may provide solutions for particular scenarios, one needs to ensure that they do not overshoot their targets, leading to a stifling of novel innovations, hindering investments into the development into novel directions as they may be perceived to be too risky  \citep{lee2018ai,wooldridge2020road,hadfield2017rules}.   
Worries  have been expressed by different organisations and academic societies that too strict policies may unnecessarily affect the benefits and societal advances that novel AI technologies may have to offer.  
Regulations affect moreover big and small tech companies differently: A highly regulated domain makes it more difficult for small new start-ups, introducing an inequality and dominance of the market by a few big players \citep{lee2018ai}. It has been emphasised that neither over-regulation nor a laissez-faire approach suffices when aiming to regulate AI technologies. In order to find a balanced answer, one clearly needs to have first an understanding of how a competitive development dynamic actually could work and how governance choices impact this dynamic, a task well-suited for dynamic systems or agent-based models.

In our recent work \citep{han2019modelling,han2022voluntary,han2020Incentive,cimpeanu2022artificial}, we examine this problem using EGT modelling, see  Figure \ref{fig}. We first developed a baseline model describing a  development competition where technologists can choose a safe (SAFE) vs risk-taking (UNSAFE) course of development \citep{han2019modelling}. Namely, it considers that to reach domain supremacy through AI in a certain domain, a number of development steps or technological advancement rounds are required \citep{han2019modelling}. In each round the technologists (or players) need to choose between one of two strategic options: to follow safety precautions (the SAFE action) or ignore safety precautions (the UNSAFE action). Because it takes more time and more effort to comply with precautionary requirements, playing SAFE is not just costlier, but implies slower development speed  too, compared to playing UNSAFE. Moreover, there is a probability that a disaster occurs if UNSAFE developments take place during this competition (see \citep{han2019modelling} for a full description). 

We demonstrate that  unconditional sanctioning will negatively influence social welfare in certain conditions of a short-term race towards domain supremacy through AI technology \citep{han2020Incentive}, leading to over-regulation of beneficial innovation (see Figure \ref{fig}B).
Since data to estimate the risk of a technology is usually limited (especially at an early stage of its development or deployment), simple sanctioning of unsafe behaviour (or reward of safe behaviour) could not fully address the issue. 

To solve this critical over-regulation dilemma in AI development, we propose an alternative approach, which is to allow technologists or race participants to voluntarily commit themselves to safe innovation procedures, signaling to others their intentions \citep{han2015synergy,nesse2001evolution}. Specifically, this  bottom-up, binding  agreement (or commitment) is established for those who want to take a safe choice,  with sanctioning applied to violators of such an agreement. 
It is shown that, by allowing race participants to freely pledge their intentions and enter (or not) in bilateral commitments to act safely and avoid risks, accepting thus to be sanctioned in case of misbehavior,  high levels of the most beneficial behaviour, for the whole, are achieved in all regions of the parameter space, see Figure \ref{fig}C.

In \citep{cimpeanu2022artificial}, we explore how  different interaction structures among race participants can alter collective choices and requirements for regulatory actions. We show when participants portray a strong diversity in terms of connections and peer-influence (as modelled through scale free networks), the conflicts that exist in homogeneous settings \citep{han2019modelling} are significantly reduced, thereby reducing the need for taking regulatory actions. 

Overall, our results are directly relevant for the design of self-organized AI governance mechanisms and regulatory policies that aim to ensure an ethical and responsible AI technology development process (such as the one proposed in the EU White Paper). 

It is noteworthy that all our above-mentioned works focused on the binary extremes of race participants' behaviour, safe or unsafe development, in an effort to focus an already expansive problem into a manageable discussion. The addition of conditional, mixed, or random strategies could provide the basis for a novel piece of work. Also, as observed with conditionally safe players in the well-mixed scenario, we envisage that these additions would show little to no effect in the early regime, with the opposite being true for the late regime, at least in homogeneous network settings. Finally, it would be important to clarify the combined effects of incentive and/or commitment mechanisms within a complex network setting.

\section{Cost-efficient Interference for Promoting Collective Behaviours in Complex Networks }
\label{Section:interference}

As discussed, the problem of promoting the evolution of collective behaviour within populations of self-regarding individuals has been intensively investigated across diverse fields of behavioural, social and computational sciences \citep{nowak:2006bo,perc2017statistical}. In most studies, a desired collective behaviour such as cooperation, coordination, trust and fairness, are assumed to emerge from the combined actions of participating individuals within the populations, without taking into account the possibility of external interference and how it can be performed in a cost-efficient way.
However, in many scenarios,  such behaviours are advocated and promoted by an exogenous decision maker, who  is not part of the system  (e.g. the United Nation  interferes in  political systems for conflict resolution or  the World Wildlife Fund organisation interferes in ecosystems to maintain biodiversity).  Thus, a new set of heuristics capable of  engineering a desired collective behaviour in a self-organised MAS is required \citep{paiva2018engineering}.  

 In our  work, we  bridged this gap by employing EGT  analysis and  ABM simulations, to study cost-efficient interference strategies  for enhancing cooperation, fairness and AI safety in the context of social games, for both well-mixed  \citep{Hanaamas2015cost,han2018cost,duong2021cost} and  networked populations (including square lattice and scale free networks) \citep{Hanijcai2018,cimpeanu2019exogenous,cimpeanu2021cost}. 
%   \begin{figure*}
% \centering
% \includegraphics[width=0.7\linewidth]{figures/figure1-new.pdf}
% \caption{\textbf{a)} In a well-mixed population (size $N = 100$), the optimal threshold of $t$ where a minimal expected cost of interference is obtained while guaranteeing at least $\omega$-frequency of cooperation, which   increases with the intensity of selection (Other parameters: $T = 2; \ R = 1; \ P = 0; \ S = -1$; $\theta = 5$);  \textbf{b)} In a  square  lattice population (size $N = 100^2$), neighborhood-based (local) interference strategies are more cost efficient in ensuring high cooperation ($\approx 100\%$)  than population-based (population cooperation level) ones. Other parameters: $T = 1.8; \ R = 1; \ P = 0; \ S = 0$ (weak PD); deterministic update was used but results are robust for stochastic update  \citep{Hanijcai2018}. }
% %\vskip -6pt
% \label{fig:panel}
% \end{figure*}

%We consider finite populations (of size $N$) of individuals who interact with each other through the one-shot Prisoner's Dilemma game (PD) \citep{nowak:2006bo}, where in each interaction two players  simultaneously choose   either to cooperate (C) or defect (D). Mutual cooperation (mutual defection) yields the reward $R$ (penalty $P$) and unilateral cooperation gives the cooperator the sucker's payoff $S$ while the defector the temptation $T$. A PD is characterized by the ordering $T > R > P > S$.

In a well-mixed population, each player interacts with all others in the population while in a structured population  the player interacts with its  immediate neighbors. A player's fitness is its averaged payoff  over all its interactions, which is then used for strategy update (through social learning) \citep{sigmund:2010aa}. Namely,  a player A with fitness $f_A$ chooses to copy the strategy of  a randomly selected player in the population (well-mixed) or  randomly selected neighbor (structured) with a probability given by the Fermi function,  $(1+e^{\beta(f_A-f_B)})^{-1}$, where $\beta$ represents the intensity of selection \citep{traulsen2006}. When $\beta = 0$  corresponds to neutral drift while $\beta \rightarrow \infty$ leads to increasingly deterministic selection. Weak or even close to neutral selections (small $\beta$) are abundant in nature, while the strong selection regime has been reported as predominant in social settings. As an alternative to this   stochastic update rule, one can also consider a  deterministic update in which agents copy, if advantageous, the most successful player in their neighbourhood.  

An \textit{interference strategy} or \textit{scheme} can be generally defined as a sequence of decisions about which players in the population to invest in  (i.e. reward the player an amount, denoted by $\theta$), in order to achieve the highest level of cooperation while minimising the total cost of investment. These decisions can be made by considering  different aspects of the population such as its global statistics and/or its structural properties. 
In the context of a well-mixed population,  an interference scheme solely depends on its composition (e.g., how many cooperators and defectors players there are at the time of decision making). 
In this case, we have derived analytical conditions for which a general interference scheme  can guarantee a given level of desired behaviour while at the same time minimising the total cost of investment (e.g., for rewarding cooperative behaviours), and show that the results are highly sensitive to the intensity of selection by interference \citep{han2018cost,duong2021cost}. 
Moreover, we have studied a specific  class of interference strategies  that make investments  whenever the number of  players with a desired behaviour (e.g., cooperative or fair players) reaches a certain threshold, denoted by $t$ ($\forall t \in \{1, \dots, N-1\}$), showing that there is a wide range of $t$ that it outperforms  standard institutional incentive strategies---which unconditionally interfere into the system regardless of its composition, corresponding to $t = N-1$ \citep{chen2015first}.  %Figure \ref{fig:panel}a  shows the optimal threshold of $t$  for varying the intensity of selection, $\beta$, where the minimal expected cost of interference is obtained while guaranteeing at least $\omega$-frequency ($0 \leq \omega \leq 1$) of cooperation. We can observe that, when $\beta$ is sufficiently small (weak selection), an intermediate value of  $t$ would lead to most cost-efficient interference strategies, while for a sufficiently strong selection, it is best to always interfere, meaning that  standard institutional incentive strategies (i.e. $t = N-1$) would be most effective. 
 
%We investigate two general classes or approaches of  interference strategies; the first is 
%based on the current composition of the population while the second is based on local neighbourhood properties, in particular, for any agent being considered for investment we examine its current cooperativeness level (the number of its neighbours who will cooperate). 
With  a structured population,  individuals (even of the same strategy) might reside in different kinds of neighborhood (with different cooperativeness levels), and therein local information might be useful to enhance cost-efficiency and cooperation.  
To this end, we test several interference paradigms  \citep{Hanijcai2018,cimpeanu2021cost,cimpeanu2019exogenous} that make investment decisions  based on a player's current  level of desirable behaviour (e.g., the number of cooperators in the neighborhood), and compared their efficiency with the population-based strategies (as in the well-mixed case).  
Furthermore, in this context, we study the impact of social  diversity, employed using heterogeneous networks of interaction (e.g. scale free networks). We show that social diversity indeed substantially influences the choice of investment approaches available to institutions. Investment is not trivial in these settings, contrary to the findings in well-mixed and lattice populations \citep{cimpeanu2021cost}. Counterintuitively, incentivising positive behaviour can lead to the exploitation of cooperators, harming pro-sociality in lieu of fostering it. We observe that highly clustered scale-free networks make it easy to select the most effective candidates for receiving endowments.

%Our systematic analysis reveals a simple strategy that invests  when there is at least one D player  in the neighborhood and does not invest otherwise,  is highly cost-efficient in promoting cooperation (see Figure \ref{fig:panel}b).
%Furthermore, when additional information regarding the fitness levels (i.e. individual income information) of  players in a neighbourhood is accessible, further improvement can be made by more accurately  influencing D neighbours for behavioural change (to become cooperators).  

The interference mechanisms we have proposed thus far have advanced the literature on external interference, but our contribution in this regard remains incipient.
Future works include analysis of  more complex interference strategies such as those vary the cost of investment over time  or combine different forms of incentives \citep{chen2015first,Han2016AAAI}. 
 We might also resort to more complex optimisation algorithms and control theory techniques, to solve the bi-objective optimisation problem posed by cost-effective interference, therefore identifying solutions that are robust across a wider range of MAS interaction settings.

 %In short, we have  studied how   cooperation can be promoted in a cost-efficient way from an external decision maker's perspective. It provides  new insights regarding  heuristics capable of  engineering a desired collective behaviour in a self-organised complex system, not only in social and biological contexts, but also Artificial Life scenarios  such as swarm-based and multi-robots systems \citep{bonabeau,key:HanetalAlife}. 

\section{Equilibrium Statistics in Random Evolutionary Games}
\label{Section:randomgames}
%\tr{TO EXPLAIN HOW RANDOM GAMES APPLY TO MAS CONTEXT}

Random evolutionary games, in which the payoff entries are random variables, have been employed extensively to model  multi-agent strategic interactions in which  very limited information about them is available, or where the environment changes so rapidly and frequently that one cannot describe the payoffs of their inhabitants' interactions  \citep{HTG12,DuongHanJMB2016,gokhale:2010pn,DuongTranHanJMB}. 
Equilibrium points of such a dynamical system are the compositions of strategy frequencies where all the strategies have the same average accumulated payoff (or fitness). They thus predict potential co-existence of different strategic behaviours in a dynamical MAS.

%Random evolutionary games in which the payoff entries are random variables form an important subclass of EGT. They are necessary to model social and biological systems in which very limited information is available, or where the environment changes so rapidly and frequently that
%one cannot describe the payoffs of their inhabitants' interactions \citep{fudenberg:1992bv, gross2009generalized}. As in classical game theory with the Nash equilibrium, see e.g. \citep{mclennan2005expected}, the analysis of properties of equilibrium points in EGT has been of special interest, see e.g. \citep{gokhale:2010pn}. These equilibrium points  predict the composition of strategy frequencies where all the strategies have the same average fitness. 

In random games, due to the randomness of the payoff entries, it is essential to study statistical properties of equilibria. How to determine the distribution of internal equilibria in random evolutionary games is an intensely investigated subject with numerous practical ramifications in diverse fields including MAS, evolutionary biology, social sciences and economics, 
providing essential understanding of complexity in a dynamical system, such as its behavioural, cultural or biological diversity and the maintenance of polymorphism. Properties of equilibrium points, particularly the probability of observing the maximal number of equilibrium points, the attainability and stability of the patterns of evolutionarily stable strategies
have been studied  \citep{gokhale:2010pn,HTG12}. However, as these prior works used a direct approach that consists of solving a system of polynomial equations, the mathematical analysis was mostly restricted to evolutionary games with a small number of players, due to the impossibility of solving general polynomial equations of a high degree (according to Abel–Ruffini's theorem).

In our work, we analyse  random evolutionary games with an arbitrary number of players \citep{DuongHanJMB2016, DuongTranHanJMB}. The key technique that we develop is to connect the number of equilibria in an evolutionary game to the number of real roots of a system of multi-variate random polynomials \citep{EK95}. Assuming that we consider $d$-player $n$-strategy evolutionary games, then the system consists of $n-1$ polynomial equations of degree $d-1$:  
\begin{equation*}
\label{eq: eqn for fitnessy}
 \sum\limits_{\substack{0\leq k_{1}, ..., k_{n-1}\leq d-1,\\  \sum\limits^{n-1}_{i = 1}k_i \leq d-1  }}\beta^i_{k_{1}, ..., k_{n-1} }\begin{pmatrix}
d-1\\
k_{1}, ..., k_{n}
\end{pmatrix} \prod\limits_{i=1}^{n-1}y_{i}^{k_i} = 0,
\end{equation*}
for $i = 1, \dots, n-1$. Here $\beta^{i}_{k_{1}, ..., k_{n-1} }:= \alpha^{i}_{k_{1}, ..., k_{n} } -\alpha^{n}_{k_{1}, ..., k_{n} }$ where $\alpha^{i_0}_{k_{1}, ..., k_{n} } := \alpha^{i_0}_{i_1,\ldots,i_{d-1}}$ is the payoff of the focal player and $k_i$, $1 \leq i \leq n$, with $\sum^{n}_{i = 1}k_i = d-1$, is the number of players using strategy $i$ in $\{i_1,\ldots,i_{d-1}\}$.
In  \citep{DuongHanJMB2016}, we analyze the mean number $E(n, d)$ and the expected density $f(n,d)$ of internal equilibria in a general $d$-player $n$-strategy evolutionary game when the individuals' payoffs are \textit{independent, normally distributed}.  We provide  computationally implementable formulas of these quantities for the general case and characterize their asymptotic behaviour for the two-strategy games (i.e. $E(2,d)$ and $f(2,d)$), estimating their  lower and upper bounds as $d$ increases. For instance, under certain assumptions on the payoffs, we obtain 
\begin{itemize}
\item Asymptotic behaviour of $E(2,d)$:
\begin{equation*}
\sqrt{d-1}\lesssim E(2,d)\lesssim \sqrt{d-1}\ln(d-1).
\end{equation*}
As a consequence, 
$
\lim\limits_{d\rightarrow\infty}\frac{\ln E(2,d)}{\ln(d-1)}=\frac{1}{2}.
$
\item Explicit formula of $E(n,2)$:
$
E(n,2)=\frac{1}{2^{n-1}}.
$
\end{itemize}
For a general $d$-player $n$-strategy game, as supported by extensive numerical results, we describe a conjecture regarding the asymptotic behaviours of $E(n,d)$ and $f(n,d)$. We also show that the probability of seeing the maximal possible number of equilibria tends to zero when $d$ or $n$ respectively goes to infinity and that the expected number of stable equilibria is bounded within a certain interval. 

In \citep{DuongTranHanDGA}, we generalize our analysis for random  evolutionary games where the payoff matrix entries are \textit{correlated} random variables. 
In social and biological contexts, correlations may arise in various scenarios particularly when there are environmental randomness and interaction uncertainty such as  when individual contributions are correlated to the surrounding contexts (e.g. due to limited resource). We establish a closed formula for the mean numbers of internal (stable) equilibria and characterize the asymptotic behaviour of this important quantity for large group sizes and study the effect of the correlation. The results show that decreasing the correlation among payoffs (namely, of a strategist for different group compositions) leads to larger mean numbers of (stable) equilibrium points, suggesting that the system or population behavioral diversity can be promoted by increasing independence of the payoff entries.

As a further development, in \citep{DuongTranHanJMB} we derive a closed formula for the distribution of internal equilibria, for both normal and uniform distributions of the game payoff entries. We also provide several universal upper and lower bound estimates,  which are independent of the underlying payoff distribution, for the probability of obtaining a certain number of internal equilibria. The distribution of equilibria provides more elaborate  information about the level of complexity or the number of different states of biodiversity that will occur in a dynamical  system, compared to what obtained with the expected number of internal equilibria.

In short, by connecting EGT to random polynomial theory, we have achieved new results on the expected number and distribution of internal equilibria in multi-player multi-strategy games. Our studies provide new insights into the overall complexity of dynamical MAS, as the numbers of players and  strategies in a MAS interaction increase. As the theory of random polynomials is rich, we expect that our novel approach can be extended to obtain results for other more complex models in population and learning dynamics such as the replicator-mutator equation and evolutionary games with environmental feedback. 
\section{Concluding Remarks}
Since its inception, research in EGT has been both exciting and challenging, as well as highly rewarding and inspiring. 
Analysing large-scale dynamical MAS of multiple interacting agents, with diverse strategic behaviours in co-presence, is highly complex and requires innovative combinations of diverse mathematical methods and simulation techniques \citep{perc2017statistical}.  On the other hand, EGT research has proven extremely powerful and found its applications in so many fields, leading to important findings  often being  reported in prestigious venues.    

Despite the many years of active research, there are still a  number of important open problems in EGT research. The problem of explaining the mechanisms of collective behaviour such as cooperation and altruism, is still far from settled \citep{pennisi2005did}. Also,  extending and generalising the existing set of mathematical techniques and modelling  tools to capture and understand realistic and ever more complex systems, is crucial. For example, modern societies become increasingly more convoluted with the advancement of technologies, changing the way  humans  live and interact with others. How EGT  can be applied to model this hybrid society and understand its dynamics is very challenging \citep{paiva2018engineering}. But solving it can prove very rewarding as it can provide insights to design appropriate mechanisms to ensure the greatest benefit for our societies, or at least to avoid existential risks that we  might otherwise have to face. 
\section{Acknowledgement}
I would like to thank my group members (both current and formal), Theodor Cimpenu, Marcus Krellner, Cedric Perret, Bianca Ogbo, Zainab Alalawi, and Aiman Elgarig,  and my collaborators, Luis Moniz Pereira, Tom Lenaerts, Francisco C. Santos, Hong Duong, Long Tran-Thanh, Alessandro Di Stefano, and Simon Powers,    who have made all the works reported here possible. 
T.A.H. is supported by Future of Life Institute AI safety grant (RFP2-154) and  Leverhulme Research Fellowship (RF-2020-603/9).
%%%%%%%%%%% The article body starts:

%\section{}\label{s1}

%\subsection{}\label{s1.1}

%\begin{figure}[t]
%\caption{Figure caption.}\label{f1}
%\end{figure}

%\begin{table*}
%\caption{} \label{t1}
%\begin{tabular}{lll}
%\hline
%&&\\
%&&\\
%\hline
%\end{tabular}
%\end{table*}

%%%%%%%%%%% The bibliography starts:

%%  ios1.bst will be used to                               %%

\bibliographystyle{apalike}
%\nocite{*}
% if your bibliography is in bibtex format, use those commands:
%\bibliographystyle{ios1}           % Style BST file.
\bibliography{bibliography}        % Bibliography file (usually '*.bib')

\begin{thebibliography}{}

\bibitem[Andras et~al., 2018]{Andras2018TrustingSystems}
Andras, P., Esterle, L., Guckert, M., Han, T.~A., Lewis, P.~R., Milanovic, K.,
  Payne, T., Perret, C., Pitt, J., Powers, S.~T., Urquhart, N., and Wells, S.
  (2018).
\newblock {Trusting Intelligent Machines: Deepening Trust Within
  Socio-Technical Systems}.
\newblock {\em IEEE Technology and Society Magazine}, 37(4):76--83.

\bibitem[Axelrod, 1984]{axelrod:1984yo}
Axelrod, R. (1984).
\newblock {\em The Evolution of Cooperation}.
\newblock Basic Books, New York.

\bibitem[Baum, 2017]{baum2017promotion}
Baum, S.~D. (2017).
\newblock On the promotion of safe and socially beneficial artificial
  intelligence.
\newblock {\em AI \& Society}, 32(4):543--551.

\bibitem[Beldad et~al., 2016]{Beldad:2016:a}
Beldad, A., Hegner, S., and Hoppen, J. (2016).
\newblock The effect of virtual sales agent {(VSA)} gender -- product gender
  congruence on product advice credibility, trust in {VSA} and online vendor,
  and purchase intention.
\newblock {\em Computers in Human Behavior}, 60:62--72.

\bibitem[Bonabeau et~al., 1999]{bonabeau}
Bonabeau, E., Dorigo, M., and Theraulaz, G. (1999).
\newblock {\em Swarm Intelligence: From Natural to Artificial Systems}.
\newblock Oxford University Press, USA.

\bibitem[Bratman, 1987]{key:bratman1987}
Bratman, M.~E. (1987).
\newblock {\em Intention, Plans, and Practical Reason}.
\newblock The David Hume Series, CSLI.

\bibitem[Castelfranchi and Falcone, 2010]{castelfranchi2010}
Castelfranchi, C. and Falcone, R. (2010).
\newblock {\em {Trust Theory: A Socio-Cognitive and Computational Model (Wiley
  Series in Agent Technology)}}.
\newblock Wiley.

\bibitem[Cave and {{\'O}h{\'E}igeartaigh}, 2018]{cave2018ai}
Cave, S. and {{\'O}h{\'E}igeartaigh}, S. (2018).
\newblock {An AI Race for Strategic Advantage: Rhetoric and Risks}.
\newblock In {\em AAAI/ACM Conference on Artificial Intelligence, Ethics and
  Society}, pages 36--40.

\bibitem[Charniak and Goldman, 1993]{key:charniak93}
Charniak, E. and Goldman, R.~P. (1993).
\newblock A {B}ayesian model of plan recognition.
\newblock {\em Artificial Intelligence}, 64(1):53--79.

\bibitem[Chen et~al., 2015]{chen2015first}
Chen, X., Sasaki, T., Br{\"a}nnstr{\"o}m, {\AA}., and Dieckmann, U. (2015).
\newblock First carrot, then stick: how the adaptive hybridization of
  incentives promotes cooperation.
\newblock {\em Journal of The Royal Society Interface}, 12(102):20140935.

\bibitem[Chopra and Singh, 2009]{Chopra09m.p.:multiagent}
Chopra, A.~K. and Singh, M.~P. (2009).
\newblock Multiagent commitment alignment.
\newblock In {\em AAMAS'2009}, pages 937--944.

\bibitem[Cimpeanu et~al., 2019]{cimpeanu2019exogenous}
Cimpeanu, T., Han, T.~A., and Santos, F.~C. (2019).
\newblock {Exogenous Rewards for Promoting Cooperation in Scale-Free Networks}.
\newblock In {\em ALIFE 2019}, pages 316--323. MIT Press.

\bibitem[Cimpeanu et~al., 2021]{cimpeanu2021cost}
Cimpeanu, T., Perret, C., and Han, T.~A. (2021).
\newblock {Cost-efficient interventions for promoting fairness in the ultimatum
  game}.
\newblock {\em Knowledge-Based Systems}, 233:107545.

\bibitem[Cimpeanu et~al., 2022]{cimpeanu2022artificial}
Cimpeanu, T., Santos, F.~C., Pereira, L.~M., Lenaerts, T., and Han, T.~A.
  (2022).
\newblock Artificial intelligence development races in heterogeneous settings.
\newblock {\em Scientific Reports}, 12(1):1--12.

\bibitem[Dhakal et~al., 2022]{DhakalRSOS2022}
Dhakal, S., Chiong, R., Chica, M., and Han, T.~A. (2022).
\newblock Evolution of cooperation and trust in an n-player social dilemma game
  with tags for migration decisions.
\newblock {\em Royal Society Open Science}, 9(5):212000.

\bibitem[Duong and Han, 2016]{DuongHanJMB2016}
Duong, M.~H. and Han, T.~A. (2016).
\newblock Analysis of the expected density of internal equilibria in random
  evolutionary multi-player multi-strategy games.
\newblock {\em Journal of Mathematical Biology}, 73(6):1727--1760.

\bibitem[Duong and Han, 2021]{duong2021cost}
Duong, M.~H. and Han, T.~A. (2021).
\newblock {Cost efficiency of institutional incentives for promoting
  cooperation in finite populations}.
\newblock {\em Proceedings of the Royal Society A}, 477(2254):20210568.

\bibitem[Duong et~al., 2018]{DuongTranHanDGA}
Duong, M.~H., Tran, H.~M., and Han, T.~A. (2018).
\newblock On the expected number of internal equilibria in random evolutionary
  games with correlated payoff matrix.
\newblock {\em Dynamic Games and Applications}.

\bibitem[Duong et~al., 2019]{DuongTranHanJMB}
Duong, M.~H., Tran, H.~M., and Han, T.~A. (2019).
\newblock On the distribution of the number of internal equilibria in random
  evolutionary games.
\newblock {\em Journal of Mathematical Biology}, 78(1):331--371.

\bibitem[Edelman and Kostlan, 1995]{EK95}
Edelman, A. and Kostlan, E. (1995).
\newblock How many zeros of a random polynomial are real?
\newblock {\em Bull. Amer. Math. Soc. (N.S.)}, 32(1):1--37.

\bibitem[{European Commission}, 2020]{EUAIwhitepaper2020}
{European Commission} (2020).
\newblock {White paper on Artificial Intelligence -- An European approach to
  excellence and trust}.
\newblock Technical report, European Commission.

\bibitem[Fehr and Gachter, 2002]{key:fehr2002}
Fehr, E. and Gachter, S. (2002).
\newblock Altruistic punishment in humans.
\newblock {\em Nature}, 415:137--140.

\bibitem[Gabriel, 2020]{gabriel2020artificial}
Gabriel, I. (2020).
\newblock Artificial intelligence, values, and alignment.
\newblock {\em Minds and machines}, 30(3):411--437.

\bibitem[Gintis, 2000]{gintis:2000bv}
Gintis, H. (2000).
\newblock {\em Game Theory Evolving}.
\newblock Princeton University Press, Princeton.

\bibitem[Gokhale and Traulsen, 2010]{gokhale:2010pn}
Gokhale, C.~S. and Traulsen, A. (2010).
\newblock Evolutionary games in the multiverse.
\newblock {\em Proc. Natl. Acad. Sci. U.S.A.}, 107(12):5500--5504.

\bibitem[Hadfield, 2017]{hadfield2017rules}
Hadfield, G.~K. (2017).
\newblock {\em Rules for a flat world: why humans invented law and how to
  reinvent it for a complex global economy}.
\newblock Oxford University Press.

\bibitem[Han et~al., 2017]{HanJaamas2016}
Han, T., Pereira, L.~M., and Lenaerts, T. (2017).
\newblock Evolution of commitment and level of participation in public goods
  games.
\newblock {\em Autonomous Agents and Multi-Agent Systems}, 31(3):561--583.

\bibitem[Han, 2013]{HanBook2013}
Han, T.~A. (2013).
\newblock {\em Intention Recognition, Commitments and Their Roles in the
  Evolution of Cooperation: From Artificial Intelligence Techniques to
  Evolutionary Game Theory Models}.
\newblock Springer.

\bibitem[Han, 2016]{Han2016AAAI}
Han, T.~A. (2016).
\newblock Emergence of social punishment and cooperation through prior
  commitments.
\newblock In {\em AAAI}, pages 2494--2500.

\bibitem[Han, 2022]{HanInterface2022}
Han, T.~A. (2022).
\newblock Institutional incentives for the evolution of committed cooperation:
  ensuring participation is as important as enhancing compliance.
\newblock {\em Journal of The Royal Society Interface}, 19(188):20220036.

\bibitem[Han et~al., 2022]{han2022voluntary}
Han, T.~A., Lenaerts, T., Santos, F.~C., and Pereira, L.~M. (2022).
\newblock Voluntary safety commitments provide an escape from over-regulation
  in ai development.
\newblock {\em Technology in Society}, 68:101843.

\bibitem[Han et~al., 2018]{Hanijcai2018}
Han, T.~A., Lynch, S., Tran-Thanh, L., and Santos, F.~C. (2018).
\newblock Fostering cooperation in structured populations through local and
  global interference strategies.
\newblock In {\em IJCAI-ECAI}, pages 289--295. AAAI Press.

\bibitem[Han and Pereira, 2013a]{han2013context}
Han, T.~A. and Pereira, L.~M. (2013a).
\newblock Context-dependent incremental decision making scrutinizing the
  intentions of others via bayesian network model construction.
\newblock {\em Intelligent Decision Technologies}, 7(4):293--317.

\bibitem[Han and Pereira, 2013b]{HanChapterBehRecog2013}
Han, T.~A. and Pereira, L.~M. (2013b).
\newblock Intention-based decision making and its applications.
\newblock In Guesgen, H. and Marsland, S., editors, {\em Human Behavior
  Recognition Technologies: Intelligent Applications for Monitoring and
  Security}, chapter Chapter 9, pages 174--211. IGI Global.

\bibitem[Han and Pereira, 2013c]{hanpereiraAICOM2013}
Han, T.~A. and Pereira, L.~M. (2013c).
\newblock State-of-the-art of intention recognition and its use in decision
  making -- a research summary.
\newblock {\em AI Communication Journal}, 26(2):237--246.

\bibitem[Han et~al., 2021a]{han2020Incentive}
Han, T.~A., Pereira, L.~M., Lenaerts, T., and Santos, F.~C. (2021a).
\newblock {Mediating Artificial Intelligence Developments through Negative and
  Positive Incentives}.
\newblock {\em PLOS ONE}, 16(1):e0244592.

\bibitem[Han et~al., 2011a]{key:hanetalAdaptiveBeh}
Han, T.~A., Pereira, L.~M., and Santos, F.~C. (2011a).
\newblock Intention recognition promotes the emergence of cooperation.
\newblock {\em Adaptive Behavior}, 19(3):264--279.

\bibitem[Han et~al., 2011b]{key:hanijcai2011}
Han, T.~A., Pereira, L.~M., and Santos, F.~C. (2011b).
\newblock The role of intention recognition in the evolution of cooperative
  behavior.
\newblock In Walsh, T., editor, {\em Proceedings of the 22nd international
  joint conference on Artificial intelligence (IJCAI'2011)}, pages 1684--1689.
  AAAI.

\bibitem[Han et~al., 2012a]{key:HanetalAlife}
Han, T.~A., Pereira, L.~M., and Santos, F.~C. (2012a).
\newblock Corpus-based intention recognition in cooperation dilemmas.
\newblock {\em Artificial Life journal}, 18(4):365--383.

\bibitem[Han et~al., 2012b]{key:Hanetall_AAMAS2012}
Han, T.~A., Pereira, L.~M., and Santos, F.~C. (2012b).
\newblock The emergence of commitments and cooperation.
\newblock In {\em AAMAS'2012}, pages 559--566.

\bibitem[Han et~al., 2013a]{han2013good}
Han, T.~A., Pereira, L.~M., Santos, F.~C., and Lenaerts, T. (2013a).
\newblock Good agreements make good friends.
\newblock {\em Scientific reports}, 3(2695).

\bibitem[Han et~al., 2013b]{ijcai2013TAH}
Han, T.~A., Pereira, L.~M., Santos, F.~C., and Lenaerts, T. (2013b).
\newblock {Why Is It So Hard to Say Sorry: The Evolution of Apology with
  Commitments in the Iterated Prisoner's Dilemma}.
\newblock In {\em IJCAI'2013}, pages 177--183. AAAI Press.

\bibitem[Han et~al., 2020]{han2019modelling}
Han, T.~A., Pereira, L.~M., Santos, F.~C., and Lenaerts, T. (2020).
\newblock {To Regulate or Not: A Social Dynamics Analysis of an Idealised AI
  Race}.
\newblock {\em Journal of Artificial Intelligence Research}, 69:881--921.

\bibitem[Han et~al., 2021b]{HanetalTrust2021}
Han, T.~A., Perret, C., and Powers, S.~T. (2021b).
\newblock When to (or not to) trust intelligent machines: Insights from an
  evolutionary game theory analysis of trust in repeated games.
\newblock {\em Cognitive Systems Research}, 68:111--124.

\bibitem[Han et~al., 2015a]{han2015synergy}
Han, T.~A., Santos, F.~C., Lenaerts, T., and Pereira, L.~M. (2015a).
\newblock Synergy between intention recognition and commitments in cooperation
  dilemmas.
\newblock {\em Scientific reports}, 5(9312).

\bibitem[Han and Tran-Thanh, 2018]{han2018cost}
Han, T.~A. and Tran-Thanh, L. (2018).
\newblock Cost-effective external interference for promoting the evolution of
  cooperation.
\newblock {\em Scientific reports}, 8.

\bibitem[Han et~al., 2015b]{Hanaamas2015cost}
Han, T.~A., Tran-Thanh, L., and Jennings, N.~R. (2015b).
\newblock The cost of interference in evolving multiagent systems.
\newblock In {\em 14th International Conference on Autonomous Agents and
  Multiagent Systems}, pages 1719--1720.

\bibitem[Han et~al., 2012c]{HTG12}
Han, T.~A., Traulsen, A., and Gokhale, C.~S. (2012c).
\newblock On equilibrium properties of evolutionary multi-player games with
  random payoff matrices.
\newblock {\em Theoretical Population Biology}, 81(4):264 -- 272.

\bibitem[Hardin, 1968]{hardin:1968mm}
Hardin, G. (1968).
\newblock The tragedy of the commons.
\newblock {\em Science}, 162:1243--1248.

\bibitem[Hasan and Raja, 2013]{hasan2013emergence}
Hasan, M.~R. and Raja, A. (2013).
\newblock Emergence of cooperation using commitments and complex network
  dynamics.
\newblock In {\em IEEE/WIC/ACM Intl Joint Conferences on Web Intelligence and
  Intelligent Agent Technologies}, pages 345--352.

\bibitem[Hauser, 2007]{ha07a-moral}
Hauser, M.~D. (2007).
\newblock {\em Moral Minds, How Nature Designed Our Universal Sense of Right
  and Wrong}.
\newblock Little Brown.

\bibitem[Hofbauer and Sigmund, 1998]{hofbauer:1998mm}
Hofbauer, J. and Sigmund, K. (1998).
\newblock {\em Evolutionary Games and Population Dynamics}.
\newblock Cambridge University Press, Cambridge.

\bibitem[Lee, 2018]{lee2018ai}
Lee, K.-F. (2018).
\newblock {\em AI superpowers: China, Silicon Valley, and the new world order}.
\newblock Houghton Mifflin Harcourt.

\bibitem[Luhmann, 1979]{Luhmann:1979:a}
Luhmann, N. (1979).
\newblock {\em Trust and {Power}}.
\newblock John Wiley \& Sons, Chichester.

\bibitem[Maynard~Smith and Price, 1973]{SP73}
Maynard~Smith, J. and Price, G.~R. (1973).
\newblock The logic of animal conflict.
\newblock {\em Nature}, 246:15--18.

\bibitem[Meltzoff, 2005]{key:Meltzoff2005}
Meltzoff, A.~N. (2005).
\newblock Imitation and other minds: the ``like me" hypothesi.
\newblock In {\em Perspectives on imitation: From neuroscience to social
  science. Imitation, human development, and culture}, pages 55--77. Cambridge,
  MA: MIT Press.

\bibitem[Nesse, 2001]{nesse2001evolution}
Nesse, R.~M. (2001).
\newblock {\em Evolution and the capacity for commitment}.
\newblock Foundation series on trust. Russell Sage.

\bibitem[Nowak, 2006a]{nowak:2006bo}
Nowak, M.~A. (2006a).
\newblock {\em Evolutionary Dynamics}.
\newblock Harvard University Press, Cambridge, MA.

\bibitem[Nowak, 2006b]{nowak:2006pw}
Nowak, M.~A. (2006b).
\newblock Five rules for the evolution of cooperation.
\newblock {\em Science}, 314:1560--1563.

\bibitem[Ogbo et~al., 2021]{ogbo2021evolution}
Ogbo, N.~B., Elragig, A., and Han, T.~A. (2021).
\newblock Evolution of coordination in pairwise and multi-player interactions
  via prior commitments.
\newblock {\em Adaptive Behavior}, page 1059712321993166.

\bibitem[O'Keefe et~al., 2020]{o2020windfall}
O'Keefe, C., Cihon, P., Garfinkel, B., Flynn, C., Leung, J., and Dafoe, A.
  (2020).
\newblock The windfall clause: Distributing the benefits of ai for the common
  good.
\newblock In {\em Proceedings of the AAAI/ACM Conference on AI, Ethics, and
  Society}, pages 327--331.

\bibitem[Paiva et~al., 2018]{paiva2018engineering}
Paiva, A., Santos, F.~P., and Santos, F.~C. (2018).
\newblock Engineering pro-sociality with autonomous agents.
\newblock In {\em Thirty-second AAAI conference on artificial intelligence}.

\bibitem[Pennisi, 2005]{pennisi2005did}
Pennisi, E. (2005).
\newblock How did cooperative behavior evolve?
\newblock {\em Science}, 309(5731):93--93.

\bibitem[Perc et~al., 2017]{perc2017statistical}
Perc, M., Jordan, J.~J., Rand, D.~G., Wang, Z., Boccaletti, S., and Szolnoki,
  A. (2017).
\newblock Statistical physics of human cooperation.
\newblock {\em Physics Reports}, 687:1--51.

\bibitem[Pereira et~al., 2021]{pereira2021employing}
Pereira, L.~M., Han, T.~A., and Lopes, A.~B. (2021).
\newblock Employing ai to better understand our morals.
\newblock {\em Entropy}, 24(1):10.

\bibitem[Pereira et~al., 2017]{Luis2017AAMAS}
Pereira, L.~M., Lenaerts, T., Martinez-Vaquero, L.~A., and Han, T.~A. (2017).
\newblock Social manifestation of guilt leads to stable cooperation in
  multi-agent systems.
\newblock In {\em AAMAS}, pages 1422--1430.

\bibitem[Pereira and Lopes, 2020]{pereira2020machine}
Pereira, L.~M. and Lopes, A.~B. (2020).
\newblock {\em Machine ethics: from machine morals to the machinery of
  morality}.
\newblock Springer.

\bibitem[Pu and Chen, 2007]{Pu:2007:a}
Pu, P. and Chen, L. (2007).
\newblock Trust-inspiring explanation interfaces for recommender systems.
\newblock {\em Knowledge-Based Systems}, 20(6):542--556.

\bibitem[Roy et~al., 2007]{key:roy2007}
Roy, P., Bouchard, B., Bouzouane, A., and Giroux, S. (2007).
\newblock A hybrid plan recognition model for alzheimer's patients:
  interleaved-erroneous dilemma.
\newblock In {\em Proceedings of IEEE/WIC/ACM International Conference on
  Intelligent Agent Technology}, pages 131--137.

\bibitem[Sadri, 2011]{key:sadri2010_logicbasedIR}
Sadri, F. (2011).
\newblock Logic-based approaches to intention recognition.
\newblock In {\em Handbook of Research on Ambient Intelligence: Trends and
  Perspectives}, pages 346--375.

\bibitem[Santos and Pacheco, 2011]{santos:2011pn}
Santos, F.~C. and Pacheco, J.~M. (2011).
\newblock Risk of collective failure provides an escape from the tragedy of the
  commons.
\newblock {\em Proc Natl Acad Sci U S A}.

\bibitem[Santos et~al., 2020]{santos2020picky}
Santos, F.~P., Mascarenhas, S., Santos, F.~C., Correia, F., Gomes, S., and
  Paiva, A. (2020).
\newblock Picky losers and carefree winners prevail in collective risk dilemmas
  with partner selection.
\newblock {\em Autonomous Agents and Multi-Agent Systems}, 34(2):1--29.

\bibitem[Sigmund, 2010]{sigmund:2010bo}
Sigmund, K. (2010).
\newblock {\em The calculus of selfishness}.
\newblock Princeton Univ. Press.

\bibitem[Sigmund et~al., 2010]{sigmund:2010aa}
Sigmund, K., De~Silva, H., Traulsen, A., and Hauert, C. (2010).
\newblock Social learning promotes institutions for governing the commons.
\newblock {\em Nature}, 466:861--863.

\bibitem[Singh, 2013]{singh2013norms}
Singh, M.~P. (2013).
\newblock Norms as a basis for governing sociotechnical systems.
\newblock {\em ACM Transactions on Intelligent Systems and Technology (TIST)},
  5(1):21.

\bibitem[Sukthankar et~al., 2014]{sukthankar2014plan}
Sukthankar, G., Geib, C., Bui, H., Pynadath, D., and Goldman, R.~P. (2014).
\newblock {\em Plan, activity, and intent recognition: Theory and practice}.
\newblock Newnes.

\bibitem[Tomasello, 2008]{key:tomassello2008}
Tomasello, M. (2008).
\newblock {\em Origins of Human Communication}.
\newblock MIT Press.

\bibitem[Traulsen et~al., 2006]{traulsen2006}
Traulsen, A., Nowak, M.~A., and Pacheco, J.~M. (2006).
\newblock Stochastic dynamics of invasion and fixation.
\newblock {\em Phys. Rev. E}, 74:11909.

\bibitem[Tuyls and Parsons, 2007]{tuyls2007evolutionary}
Tuyls, K. and Parsons, S. (2007).
\newblock What evolutionary game theory tells us about multiagent learning.
\newblock {\em Artificial Intelligence}, 171(7):406--416.

\bibitem[Von~Neumann and Morgenstern, 1944]{von2007theory}
Von~Neumann, J. and Morgenstern, O. (1944).
\newblock {\em Theory of games and economic behavior}.
\newblock Princeton university press.

\bibitem[Wooldridge, 2020]{wooldridge2020road}
Wooldridge, M. (2020).
\newblock {\em The road to conscious machines: The story of AI}.
\newblock Penguin UK.

\bibitem[Wooldridge and Jennings, 1999]{Wooldridge99}
Wooldridge, M. and Jennings, N.~R. (1999).
\newblock The cooperative problem-solving process.
\newblock In {\em Journal of Logic and Computation}, pages 403--417.

\end{thebibliography}

% or include bibliography directly:
%\begin{thebibliography}{0}
%\bibitem{r1} F. Author, Information about cited object.
%
%\bibitem{r2} S. Author and T. Author, Information about cited object.
%\end{thebibliography}

\end{document}